\documentclass{article}
\usepackage{ijcai21}

\usepackage{times}

\usepackage{soul}
\usepackage[hyphens]{url}
\usepackage[hidelinks]{hyperref}
\usepackage[utf8]{inputenc}
\usepackage[small]{caption}
\usepackage{graphicx}
\usepackage{amsmath}
\usepackage{booktabs}
\usepackage{todonotes}
\usepackage[normalem]{ulem}
\usepackage{color}

\usepackage{soul} 
\title{Automated Fact-Checking for Assisting Human Fact-Checkers}

\author{%
Preslav Nakov$^1$\footnote{Contact Author}\and
David Corney$^2$ \and
Maram Hasanain$^{3}$\and
Firoj Alam$^1$\and
Tamer Elsayed$^3$\and   \\
Alberto Barr\'on-Cede\~no$^4$\and
Paolo Papotti$^5$\and
Shaden Shaar$^1$\and
Giovanni Da San Martino$^6$
\affiliations
$^1$Qatar Computing Research Institute, HBKU, Qatar, $^2$Full Fact, UK, $^3$Qatar University, Qatar, \\
$^4$Universit\`a di Bologna, Italy, 
$^5$EURECOM, France, $^6$University of Padova, Italy\\
\emails
\{pnakov, fialam, sshaar\}@hbku.edu.qa, david.corney@fullfact.org, a.barron@unibo.it, \{maram.hasanain, telsayed\}@qu.edu.qa, 
papotti@eurecom.fr, dasan@math.unipd.it
}

\begin{document}

\maketitle

\begin{abstract}
The reporting and the analysis of current events around the globe has expanded from professional, editor-lead journalism all the way to citizen journalism. Nowadays, politicians and other key players enjoy direct access to their audiences through social media, bypassing the filters of official cables or traditional media. However, the multiple advantages of free speech and direct communication are dimmed by the misuse of media to spread inaccurate or misleading claims. These phenomena have led to the modern incarnation of the \textit{fact-checker} --- a professional whose main aim is to examine claims using available evidence and to assess their veracity. As in other text forensics tasks, the amount of information available makes the work of the fact-checker more difficult. 
With this in mind, starting from the perspective of the professional fact-checker, we survey the available intelligent technologies that can support the human expert in the different steps of her fact-checking endeavor. These include identifying claims worth fact-checking, detecting relevant previously fact-checked claims, retrieving relevant evidence to fact-check a claim, and actually verifying a claim. In each case, we pay attention to the challenges in future work and the potential impact on real-world fact-checking.
\end{abstract}

\section{Introduction}

The spread of fake news, misinformation and disinformation on the web and in social media has become an urgent social and political issue. Social media have been widely used not only for social good, but also to mislead entire communities. To fight against such false or misleading information, several initiatives for manual   fact-checking have been launched. Some notable fact-checking organizations include \emph{FactCheck.org},\footnote{\label{factcheck}\url{http://www.factcheck.org}}
\emph{Snopes},\footnote{\label{snopes}\url{http://www.snopes.com/fact-check/}}
\emph{PolitiFact},\footnote{\label{politifact}\url{http://www.politifact.com}}
and \emph{FullFact}.\footnote{\label{fullfact}\url{http://fullfact.org}} 

Such fact-checking organizations are also potential beneficiaries of and/or leaders in automated fact-checking research. As misinformation and disinformation have become major concerns globally, tech companies, as well as national and international agencies began work in this area. Recently, several international initiatives have also emerged such as the \textit{Credibility Coalition}\footnote{\label{credibilitycoalition}\url{https://credibilitycoalition.org}} and \textit{EUfactcheck},\footnote{\label{eufactcheck}\url{https://eufactcheck.eu}} and some tools have been made available such as \emph{Google Factcheck}\footnote{\label{google-toolbox}\url{https://toolbox.google.com/factcheck/explorer}} and \emph{Hoaxy}.\footnote{\label{hoaxy}\url{https://hoaxy.osome.iu.edu}} Moreover, fact-checking is a common task in settings that go beyond online misinformation, as the verification of content's accuracy is a priority for many organizations~\cite{KaragiannisSPT20}.

A large body of research has been devoted to developing automatic systems for fact-checking \cite{Li:2016:STD:2897350.2897352,Shu:2017:FND:3137597.3137600,Lazer1094,Vosoughi1146,vo2018rise}. This includes datasets \cite{Hassan:15,augenstein-etal-2019-multifc}, and evaluation campaigns \cite{thorne-vlachos:2018:C18-1,ECIR2021:CLEF}. However, there are credibility issues with automated systems~\cite{fullfact:coof}, and thus a reasonable solution (i.e.,~human in the loop) is to facilitate human fact-checkers using automated systems. Yet, there has been limited work in this direction.

Thus, to facilitate human fact-checkers, in this survey, we explore what fact-checkers want and what research has been done that can actually support them in their work. This is important because manual fact-checking is a time-consuming process, going through several manual steps. The study by \citeauthor{vlachos-riedel-2014-fact}~\shortcite{vlachos-riedel-2014-fact} describes the following typical sequence of fact-checking steps: (\emph{i})~extracting statements that are to be fact-checked, (\emph{ii})~constructing appropriate questions, (\emph{iii})~obtaining the pieces of evidence from relevant sources, and (\emph{iv})~reaching a verdict using that evidence. 

In the current information ecosystem (including web and social media), there is a large volume of false claims not only in textual form, but also misleading or manipulated images and videos, including ``deepfakes,'' and there has been a lot of recent work on fact-checking images and videos. However, here we limit our focus to automated fact-checking on text, as it remains the focus of most professional fact-checkers.

There have been a number of surveys on ``fake news'' \cite{Shu:2017:FND:3137597.3137600,Lazer1094,Vosoughi1146,Survey:2021:Multimodal:Disinformation}, rumors \cite{10.1145/3161603}, fact-checking \cite{thorne-vlachos:2018:C18-1,Kotonya2020}, factuality \cite{Li:2016:STD:2897350.2897352,DBLP:journals/jdiq/ZannettouSBK19,Survey:2021:Media:Factuality:Bias}, and propaganda \cite{da2020survey}. Unlike that work, here we study the desiderata of fact-checkers vs. the research attempts that aim to meet them.

\section{What Fact-Checkers Want}

Recently, Full Fact carried out extensive interviews with professional fact-checkers from 24 organizations in 50 countries~\cite{fullfact:coof}. The report discussed key challenges they face where they believe technology can help. 
These include monitoring potentially harmful content, selecting claims to check, creating and distributing articles, and managing suggestions from readers (such as tip lines serving \mbox{WhatsApp} or Signal).

The same report revealed that most fact-checkers do \emph{not} believe that tools to automate the verification of claims, i.e.,~the last step of a typical fact-checking pipeline~\cite{vlachos-riedel-2014-fact}, will be used in the foreseeable future. Some believe that the required intuition and creativity can never be automated, even if some parts of their work can be supported.

This sets up a twin challenge for Artificial Intelligence (AI) practitioners: \textit{first}, to develop practical tools that solve the problems fact-checkers face, and \textit{second}, to demonstrate their value to fact-checkers in their day-to-day work. In the meantime, there is a recognised need for tools to help with finding claims, including previously fact-checked claims, and in providing relevant evidence to help write fact-checking articles.

\subsection{Finding Claims Worth Fact-Checking}

Choosing which claims to check is a complex process. \mbox{Fact-checking} is time-consuming and it often takes effort to determine whether a claim can even be checked, let alone whether it is misleading. Fact-checkers have to balance the potential harm that a misleading claim may cause (including risk to health, risk to democratic processes, and risk of exacerbating emergency situations) against the effort required to check a claim. Fact-checkers are also committed to being non-partisan, and thus it is important that such tools do not introduce any unfair bias. In many countries, governments choose not to publish reliable official statistics, thus making certain statistics-related claims virtually impossible to verify. 

While simple algorithms can often decide whether content is viral, it is much harder to estimate the ``checkworthiness'' of a claim. For example, breaking news stories are often both popular and accurate. Given the limited resources of fact-checking organizations, many claims that are check-worthy nonetheless remain unchecked; thus, using historic lists of claims that were or were not checked is \emph{not} a reliable indication of whether similar claims are worth fact-checking.

Claims may be found in many sources, including news websites, social media (text, audio, or video), and broadcast media. To monitor such a range of sources, fact-checkers often use a variety of technologies, such as news alerts, automatic speech recognition and translation tools, all of which typically depend on underlying AI technologies.

\subsection{Detecting Previously Fact-Checked Claims}

Misleading claims are often repeated in multiple channels, independently of any fact-checks or rebuttals.\footnote{President Donald Trump repeated one false claim over 80 times:\\ \url{http://tinyurl.com/yblcb5q5}.} Once a claim has been established as misleading, the ongoing spread of repeats or copies of the claim can be minimised by its rapid detection. In the simplest cases, these could be simple ``copy and paste'' repeats that are relatively easy to detect, but more often they will be paraphrases of the original or endlessly evolving variations. Given the resources required to write fact-checking articles, it is preferable to respond to multiple repeats of a claim with a single fact-checking article.

The number of fact-checking initiatives continues to grow. The \emph{Duke Reporters' Lab} lists 305 active fact-checking organizations.\footnote{\url{http://reporterslab.org/fact-checking/}}
While some of them have debunked just a couple of hundred claims, others such as \emph{PolitiFact},  \emph{FactCheck.org}, \emph{Snopes}, and \emph{Full Fact} have each fact-checked thousands or even tens of thousands of claims.

Moreover, manual fact-checking often comes too late. It has been shown that ``fake news'' can spread six times faster than real ones~\cite{Vosoughi1146}, and that over half of the spread of some viral claims happens within the first ten minutes of their posting on social media~\cite{zaman2014}. To counter this, quickly detecting that a new viral claim has already been fact-checked allows for a timely action that can limit the spread and the potential harmful impact.
The problem is made harder by the transient nature of many claims. For example, a claim about infection rates may be wrong today but correct next week, and thus re-using previous checks should be done carefully.

For journalists, the ability to discover quickly whether a claim has been previously fact-checked could be revolutionizing as it would allow them to put politicians on the spot during live events. In such a scenario, automatic fact-checking would be of limited utility as, given the current state of technology, it is not credible enough in the eyes of a journalist.

Finally, false claims often originate in one language and then get translated to other languages. Tools that can spot repeated claims across languages would be useful to address this. More generally, multi-lingual tools can help fact-checkers around the world, even those with limited resources.

\subsection{Evidence Retrieval}
\label{sub:want-evidence_retrieval}

Fact-checking is often limited by the time available: there are typically far more claims to verify than what is practically possible. Even if full automation remains out of reach (see the next section), tools that support fact-checkers in their manual verification process are to be welcomed. 

Tools that automatically retrieve relevant data from trusted sources may save fact-checkers
a lot of time. This is especially true if the evidence is hidden in large text documents, audio-visual recordings and streams, or is in a language that the fact-checker is not familiar with. Thus, combining automatic transcription, summarization, translation, and search can make sources of evidence available to fact-checkers that would be impossible or impractical to access otherwise.

\subsection{Automated Verification}

On first consideration, the automated verification of claims seems like the ultimate application of AI to fact-checking. If such technologies can be developed and deployed, they would allow fact-checking organizations to be faster and to provide a more comprehensive coverage than manual fact-checking could ever achieve. However, many claims are not simply \emph{correct} or \emph{incorrect}, but may be \emph{partially correct}, or \emph{correct but misleading} without extra context, etc. One key role of professional fact-checkers is to help their audience gain full understanding of a claim, with all its nuances and complexity, rather than simply applying a binary classification.

Fact-checkers can only have an impact if they are trusted by their readers. They therefore take great care to only publish fact-checks after meticulous research, and adhere to strict editorial standards, as outlined, e.g.,~in the fact-checkers' code of principles\footnote{\url{http://www.poynter.org/ifcn-fact-checkers-code-of-principles/}} by the \emph{International Fact-Checking Network}.
This leads to a major hurdle before adopting fully automated verification methods: such methods will inevitably be imperfect, and publishing incorrect fact-checks could seriously damage the reputation of the responsible fact-checking organization. They may be more valuable as internal tools by presenting the evidence, reasoning and conclusion regarding a claim, before the (human) fact-checker writes and publishes their fact-checking article.

\section{What Technology Currently Offers}

\begin{figure}[t]
\centering
\includegraphics[width=\columnwidth]{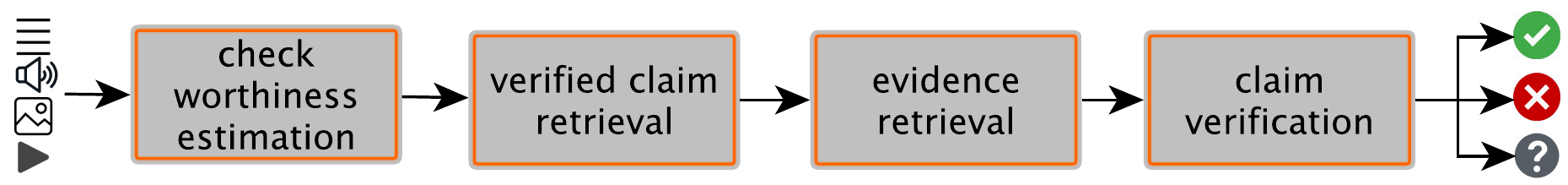}
\caption{A fact-checking pipeline.}
\label{fig:clef-pipeline}
\end{figure}

Fact-checking is not a straightforward or routine process. It requires a chain of steps that go from sensing media and spotting check-worthy claims all the way through to concluding whether the claim is true, partially-true, false, misleading, or perhaps impossible to judge. Figure~\ref{fig:clef-pipeline} shows a typical fact-checking pipeline, partially derived from~\cite{ECIR2020:CLEF}. Below, we discuss each step in this pipeline.

\subsection{Finding Claims Worth Fact-Checking}

As fact-checkers are flooded with claims, they need to decide what is actually worth fact-checking. This has encouraged the development of AI solutions, e.g.,~as part of shared tasks such as the \emph{CLEF CheckThat! lab} 2018-2021~\cite{clef2018,CheckThat:ECIR2019,ECIR2020:CLEF,ECIR2021:CLEF}, as well as inside dedicated fact-checking organizations such as Full Fact~\cite{corney:manifestos}. 

The problem is widely tackled as a ranking one, where the system has to produce a ranked list of claims coupled with check-worthiness scores. Such a score is important to increase the system's transparency and to provide fact-checkers with the ability to prioritize or to filter claims. Fact-checkers can also provide feedback on how reflective this score is of the actual check-worthiness of a claim, which can be later used to tune the system. 

\emph{ClaimBuster}~\cite{hassan2017toward} is the first system for check-worthiness detection, and it was used by fact-checkers in the \emph{Duke Reporters' Lab} project.\footnote{http://reporterslab.org/tech-and-check} It was trained on a manually annotated dataset to distinguish between \emph{non-factual sentences}, \emph{unimportant factual claims}, and \emph{check-worthy factual claims};
it used features based on sentiment, named entities, part-of-speech tags, words, and claim length. \citeauthor{konstantinovskiy2018towards}~\shortcite{konstantinovskiy2018towards} developed a more detailed schema and dataset for check-worthiness annotation of TV shows. \citeauthor{gencheva-EtAl:2017:RANLP}~\shortcite{gencheva-EtAl:2017:RANLP} created a dataset of political debates, derived by observing which sentences were fact-checked by fact-checkers; they used a rich set of features modeling the sentence structure and the context of the claim. The dataset was used in the \emph{ClaimRank} system \cite{NAACL2018:claimrank}, and was extended to multitask learning from nine fact-checking organizations \cite{RANLP2019:checkworthiness:multitask}. Further extensions were used for the \emph{CLEF CheckThat! lab}, where the participants developed models based on pre-trained transformers such as BERT and RoBERTa~\cite{clef-checkthat-Hasanain:2020,clef-checkthat-Nikolov:2020,clef-checkthat-williams:2020}. Finally, as observation-based annotations cannot give reliable negative examples, the task was also modeled using positive unlabeled learning \cite{wright-augenstein-2020-claim}.

During a recent general election, \emph{Full Fact} used a fine-tuned BERT model to classify claims made by each political party, according to whether they were \textit{numerical claims}, \textit{predictions}, \textit{personal beliefs}, etc.~\cite{corney:manifestos} This allowed fact-checkers to rapidly identify the check-worthy claims, and thus to focus their efforts in the limited time available while voters are making their final decisions. 

Social media companies are also working on combating misinformation and disinformation on their platforms. \emph{Facebook} described a proprietary tool to identify claims that should be fact-checked~\cite{facebook_fact}. They leverage flags by the users for a post indicating that it is potentially false, as well as features from the content of the replies, to predict whether the post contains false information. The model is updated using feedback from fact-checkers.

\subsection{Detecting Previously Fact-Checked Claims}

Interestingly, despite the importance of detecting whether a claim has been fact-checked before, it has been explored only recently. \citeauthor{shaar:knownlie_2020}~\shortcite{shaar:knownlie_2020} formulated the task, and released two specialized datasets: (a)~on tweets, which are to be compared to claims in \emph{Snopes}, and (b)~on political debates, to be matched to claims in \emph{PolitiFact}. 
They further proposed a learning-to-rank approach based on a combination of BERT and traditional BM25, matching the input to the entire fact-checking article. Follow-up work explored the role of context for (b), including using neighboring sentences, co-reference resolution, and reasoning over the target text with Transformer-XH \cite{previously:factchecked:context}.
The task was also featured in the \emph{CLEF CheckThat! Lab} \cite{ECIR2020:CLEF,ECIR2021:CLEF}.  \citeauthor{vo-lee-2020-facts}~\shortcite{vo-lee-2020-facts} explored a multi-modal setup, where tweets with claims about images were matched against the \emph{Fauxtography} section of \emph{Snopes}. \emph{Full Fact} is currently trialling a similar tool internally.

Recently, \emph{Google} has released the \emph{Fact Check Explorer},\textsuperscript{\ref{google-toolbox}}
which is an exploration tool that allows users to search a number of fact-checking websites, such that use \emph{ClaimReview} from \texttt{schema.org},\footnote{\url{http://schema.org/ClaimReview}} for the mentions of a topic, a person, etc. However, the tool cannot handle complex claims, as it uses \emph{Google Search}, which is not optimised for long queries.

\subsection{Evidence Retrieval}

Evidence retrieval aims to find external evidence to help fact-checkers decide on the factuality of an input claim. When the input consists of a check-worthy claim and a (potentially closed) data collection, the process could finish in the production of a ranking of the relevant data ---as in a \textit{standard} retrieval scenario--- or in the extraction of specific pieces of evidence, e.g.,~a text snippet or a recording.

When dealing with a closed reference collection, the task can be addressed as a ranking problem, e.g.,~based on BM25 or on some kind of similarity over vectorial representations between the input claim and the documents in the collection. Recent work has also combined document-level and sentence-level similarity to improve relevant document retrieval~\cite{akkalyoncu-yilmaz-etal-2019-cross}. 

Once a relevant document has been found, it is possible to further extract relevant snippets representing arguments in favour or against the target claim, to be presented to the human fact-checker~\cite{Alshomary:20}.

It is also possible to further generate snippets to brief the fact-checkers with some relevant background knowledge about the target claim. \citeauthor{fan-etal-2020-generating}~\shortcite{fan-etal-2020-generating} achieved this by first generating and retrieving relevant \emph{passage briefs}, then identifying and retrieving documents based on \emph{entity briefs}, and finally generating and answering \emph{question answering briefs} decomposed from the claim. 

The \emph{CLEF-2013 INEX lab}~\cite{Bellot:13} included a shared task that asked to retrieve evidence snippets from a pool of 50k books to \textit{confirm} or to \textit{refute} a claim. They found that entity matching was one of the most important features.

The \emph{CLEF CheckThat! lab} also featured tasks on claim evidence retrieval, at the document and also at the passage level, which was offered in Arabic~\cite{CheckThat:ECIR2019,ECIR2020:CLEF}.

The \emph{Fact Extraction and Verification shared task}~(\emph{FEVER}) focused on extracting an evidence sentence related to a claim from Wikipedia articles and determining whether it \emph{supports}, \emph{refutes}, or \emph{provides no enough information} about the claim \cite{thorne-etal-2018-fact}. As in \emph{INEX}, named entities were among the key pieces of information, and they were often used to compose the queries to retrieve the most relevant articles~\cite{malon-2018-team,hanselowski-etal-2018-ukp}. A typical system to solve the task starts with document retrieval, e.g.,~using BM25, followed by sentence retrieval based on the similarity between the input claim and each sentence in the top-n retrieved documents, which can be measured using TF.IDF, Word Mover's Distance, or BERT. Finally, it would use natural language inference to decide on the verdict. More recent work has used specialized neural semantic matching networks for each of these steps~\cite{nie2019combining}.

Evidence retrieval might need to go beyond text, e.g.,~when verifying a claim about an image or a video. In such cases, reverse image search can help find other contexts where the multimedia content was used \cite{EMNLP2019:fauxtography}. This allows to check whether these contexts agree with the claim, and to detect out-of-context content, e.g.,~an image or a video from one event portrayed as being from a different event, as well as potentially manipulated images/videos. Popular tools for this include \emph{TinEye},\footnote{\url{http://tineye.com/}} \emph{Google Image Search}, and \emph{Yandex Image Search}. Relevant research tools are also being developed in two EU projects: \emph{WeVerify}\footnote{\url{http://weverify.eu/tools/}} and \emph{InVID}.\footnote{\url{http://www.invid-project.eu}}

\subsection{Automated Verification}

Automatic claim verification approaches can be divided into explainable and non-explainable.

\emph{Explainable approaches}, also known as \emph{reference-based approaches}, are more relevant to assisting human fact-checkers. They verify the input claim against a trusted source such as tables \cite{ChenWCZWLZW20} or a database~\cite{AhmadiLPS19}, or using inference over a knowledge graph, possibly while also using Horn rules~\cite{Gad-Elrab:2019:TTF:3308558.3314126}. This includes two approaches that we discussed above: finding previously fact-checked claims that can verify the input claim~\cite{shaar:knownlie_2020}, and fact-checking it against Wikipedia~\cite{thorne-etal-2018-fact,nie2019combining}.

\emph{Non-explainable approaches} make a prediction based on the content of documents retrieved from the Web~\cite{popat2016credibility,karadzhov-etal-2017-fully,augenstein-etal-2019-multifc}, or on social media by modeling the message and its propagation, the users and their reactions over time, links to media sites, etc. \cite{10.1145/1963405.1963500,Shu:2017:FND:3137597.3137600,Vosoughi1146,FANG:2020}.
This further includes analysis of the language used in the claims based on lexicons such as LIWC~\cite{rashkin-etal-2017-truth}, or using perplexity analysis \cite{Madotto:2021:perplexity}. Fact-checking has also been done using masking in BERT-style transformers \cite{lee-etal-2020-language}. 

While automatic verification is hard, there are promising results for certain kinds of claims.
For example, an explicit claim about a numerical value, such as ``\emph{In 2017, global electricity demand grew by 3\%.}'', can be verified automatically using official statistics, even when this requires applying a complex formula~\cite{KaragiannisSPT20}. Success here depends on the availability of reliable data, presented in a consistent format, which varies widely between countries and fields.  
Similarly, simple claims can be verified with promising accuracy when good evidence is available, e.g.,~for popular entities on the Web~\cite{augenstein-etal-2019-multifc}. 

While the accuracy and the scope of automated fact-checking algorithms keeps improving, two problems prevent their adoption in fact-checking organizations. First, even on the original datasets, their effectiveness is not high enough to allow automatic decisions. Second, most claims in the public realm are more complex, e.g.,~that COVID-19 vaccines have been developed too quickly and are still experimental.\footnote{\url{http://fullfact.org/online/covid-19-survival-rate-less-998/}}

To verify such claims, fact-checkers might need to interview experts, to collaborate with other fact-checkers, to understand the context and the framing of the claims, to track down and to verify multiple sources and pieces of evidence --- all of which require human-levels intelligence. The general verification of arbitrary claims requires deep understanding of the real world that currently eludes AI. Indeed, most methods are designed to assist fact-checkers in their work with suggestions and assume that a human user will assess the verification output before assigning a true/false label.

\subsection{Some Real-World Systems}

Below, we present a brief overview of some notable systems that cover multiple steps of the fact-checking pipeline, while also offering a suitable user interface.

\paragraph{AFCNR:} The system accepts a claim as an input, searches over news articles, retrieves potential evidence and presents to the user a judgment on the stance of each piece of evidence towards the claim and an overall rating of the claim's veracity given the evidence \cite{miranda2019automated}. The system was extensively tested by eleven journalists from BBC.

\paragraph{BRENDA:} This is a browser extension,
which allows users to fact-check claims directly while reading news articles~\cite{botnevik2020brenda}. It supports two types of input, either the full page opened in the browser, or a highlighted snippet inside the page. In the first scenario, the system applies check-worthiness identification in order to decide which sentences in a page to fact-check.  

\paragraph{ClaimPortal:}\footnote{\url{http://idir.uta.edu/claimportal/}}
After retrieving tweets in response to a query, the system~\cite{majithia2019claimportal} scores them for check-worthiness using \emph{ClaimBuster} and tries to verify each tweet using previously fact-checked claims from \emph{PolitiFact}.

\paragraph{Squash:}
The system is developed at the \emph{Duke Reporters' lab}, this system (\emph{i})~listens to speech, debate and other events, (\emph{ii})~transcribes them into text, (\emph{iii})~identifies claims to check, and then (\emph{iv})~fact-check them by finding matching claims already fact-checked by humans~\cite{squash}. 

\paragraph{Full Fact's} system is designed to support fact-checkers. It (\emph{i})~follows news sites and social media, (\emph{ii})~identifies and categorizes claims in the stream, (\emph{iii})~checks whether a claim has been already verified, and then (\emph{iv})~enriches the claims with data to support the fact-checker. It is in daily use in the UK and several countries in Africa~\cite{dudfield:scaling}.

\vspace{6pt}
We believe that the prototypes presented above are good examples of the steps taken towards developing systems that cater to fact-checkers. More systems are now designed to \emph{efficiently} identify claims originating from \emph{various types} of sources (e.g.,~news articles, broadcast, and social media). Moreover, the fact-checker is now becoming a part of the system by providing feedback, rather than just being a consumer of its output. 
Finally, we see an increase in systems' transparency by providing explainable decisions, thus making them more an assistive tool rather than a replacement for the fact-checker. However, there are several challenges left to tackle, as we present in the next sections.

\section{Lessons Learned}

The main lesson from our analysis is that there is a partial disconnection between what fact-checkers want and what technology has to offer. We provide more detail below.

\begin{enumerate}
    \item Over time, many tools have been developed, either to automatically fact-check claims or to provide facilities to the fact-checkers to support their manual fact-checking 
    process. However, there are still limitations in both automated and manual processes: (\emph{i})~credibility issue for automated systems, as they do not provide supporting evidence, and (\emph{ii})~scalability issue for manual fact-checking. 
    \item Automated fact-checking systems can help fact-checkers in different ways: (\emph{i})~to find claims worth fact-checking, (\emph{ii})~to find relevant previously fact-checked claims; (\emph{iii})~to find supporting evidence (in the form of text, audio or video), translating (for multilingual content) and summarising relevant posts, articles and documents if needed, and (\emph{iv})~to detect claims that are spreading faster to slow them down. 
    
    \item There is a lack of collaboration between researchers and practitioners in terms of defining tasks and developing datasets to develop automated systems. In general, a human-in-the-loop can be an ideal setting for fact-checking, which is currently not fully explored.  

\end{enumerate}

\section{Challenges and Future Forecasting}

Below we discuss some major challenges and we forecast some promising research directions:

\subsection{Major Challenges}

\begin{itemize}
    \item \textbf{Leveraging multi-lingual resources:} The same claim, with slightly different variants, often spreads over different regions of the world at almost the same or at different time periods. These may be ``international claims'' such as medical claims about COVID-19, or stories that are presented as local, but with varied, false locations. Those claims might be fact-checked in one language, but not in others. Moreover, resources in English are abundant, but in low-resource languages, such as Arabic, they are clearly lacking. Aligning and coordinating the verification resources and leveraging them across different languages to improve fact-checking is a challenge. 
    
    \item \textbf{Ambiguity in the claims:} Another reason why automatic fact-checking is challenging is related to the fact that often a claim has multiple interpretations. An example is ``\emph{The COVID death rate is rising.}'' Is this about mortality or about fatality rate? Does it refer to today/yesterday or to the last week/month? Does it refer to the entire world or to a specific area? In such cases, knowledge about the context is necessary in order to properly frame the claim and to filter out unlikely interpretations. After that, all remaining interpretations should be analyzed, which would further slow down the work of fact-checkers. One system that proposes a solution to this problem is CoronaCheck.\footnote{\url{http://coronacheck.eurecom.fr}} 
    \item \textbf{System bias:} The majority of existing systems are trained using datasets curated by a small group of people and often annotated by non-experts. This in turn results in systems biased towards how the system developers perceive factuality and how the annotation task was described to the annotators. The dangers of bias in large language models is becoming increasingly obvious~\cite{bender2021dangers}, and should not be ignored just because the purpose of the system is benevolent.

    \item \textbf{Contextual information:} The current state-of-the-art for automated fact-checking makes limited use of contextual information, e.g.,~reader's comments, linked sources of news articles, social network data for social media posts. Such information can provide useful signals for enriching the current models.

    \item \textbf{Multimodality:} Information is typically disseminated through multiple modalities such as text, image, speech, video, temporal, user profile, and network structure. Addressing the problem based on a single modality can be a step towards failure. For example, it might be difficult to detect fake news pieces that are automatically generated using deep fakes and/or GPT-3-style text generation. To avoid such issues, multimodal approaches would be one way to go, if evidence can be gathered from multiple types of sources at the same time. This in turn requires multimodal datasets to develop suitable models.

\end{itemize}

\subsection{Future Forecasting}

\begin{itemize}
    \item \textbf{Close collaboration between fact-checking platforms and researchers:} We envision closer collaboration between professionals from fact-checking platforms alongside researchers in the domain to discuss common interests, existing solutions, and future directions, has been a challenge. 

    \item \textbf{Integrated solutions:} 
    We also envision unified and open-source initiatives to develop resources for system development and benchmarking.

    \item \textbf{Usability:} We further forecast more research on the system interface design, which would facilitate the adoption of AI by fact-checkers. It is important to develop systems that require minimal technical knowledge and reduce cognitive load. Such systems can help a larger number of fact-checkers and journalists in the fact-checking process. 
    
    \item \textbf{Interpretability and explainability:}
    Models should be designed in such a way that their outcomes are explainable, unbiased, and more accountable to ethical considerations.

    \item \textbf{Efficient and real-time solutions:} Finally, in order to tackle the velocity of the spread of fake news there is a need to develop systems that are efficient and scalable for real-time solution. To be effective, such systems would need to be embedded within, or accessible by, social networks and other big technology companies. 
\end{itemize}

\section{Conclusion}

We have presented a survey of the available intelligent technologies that can support the human experts in the different steps of the manual process of fact-checking claims. These include tasks such as identifying claims worth fact-checking, detecting relevant previously fact-checked claims, retrieving relevant evidence to support the manual fact-check of a claim, and actually verifying a claim. In each case, we paid attention to the challenges in future work and to the potential impact on real-world fact-checking.

We argued that there is currently only a partial overlap between what fact-checkers want and what the research community considers as a priority. We then discussed lessons learned and major challenges that need to be overcome. Finally, we suggested several research directions, which we forecast will emerge in the near future.

\section*{Acknowledgments}
This work was made possible in part by grant\# NPRP 7-1330-2-483 from the Qatar National Research Fund (a member of Qatar Foundation). Partial support also comes from a Google gift. This work is also part of the Tanbih mega-project, developed at the Qatar Computing Research Institute, HBKU, which aims to limit the impact of ``fake news,'' propaganda, and media bias by making users aware of what they are reading.
The statements made herein are solely the responsibility of the authors. 

\bibliographystyle{named}
\bibliography{ijcai21,clef20_checkthat}

\end{document}